\documentclass[runningheads]{llncs}

\usepackage{eccv}

\usepackage{eccvabbrv}

\usepackage{graphicx}
\usepackage{booktabs}

\usepackage[accsupp]{axessibility}

\usepackage{hyperref}

\usepackage{orcidlink}

\usepackage{pifont}
\usepackage{multirow}

\begin{document}

\newcommand{\tech}{FADA}
\newcommand{\techfull}{Frame-level Attention Distribution Alignment}

\newcommand{\methodname}{SegVGGT}
\title{\methodname: Joint 3D Reconstruction and Instance Segmentation from Multi-View Images}

\titlerunning{SegVGGT}

\author{
    Jinyuan Qu\inst{1,3}$^{*}$ \and
    Hongyang Li\inst{2,3}$^{*}$ \and
    Lei Zhang\inst{3}$^{\dagger}$
}

\authorrunning{J.~Qu et al.}

\institute{
    Tsinghua University \and
    South China University of Technology \and
    International Digital Economy Academy (IDEA)
    }

\maketitle

\begingroup
\renewcommand{\thefootnote}{\fnsymbol{footnote}}
\footnotetext[1]{Equal contribution. Work done during an internship at IDEA Research.}
\footnotetext[4]{Corresponding author.}
\setcounter{footnote}{0}
\endgroup

\begin{abstract}

\vspace{-3mm}

3D instance segmentation methods typically rely on high-quality point clouds or posed RGB-D scans, requiring complex multi-stage processing pipelines, and are highly sensitive to reconstruction noise.
While recent feed-forward transformers have revolutionized multi-view 3D reconstruction, they remain decoupled from high-level semantic understanding.
In this work, we present {\methodname}, a unified end-to-end framework that simultaneously performs feed-forward 3D reconstruction and instance segmentation directly from multi-view RGB images.
By introducing object queries that interact with multi-level geometric features, our method deeply integrates instance identification into the visual geometry grounded transformer.
To address the severe attention dispersion problem caused by the massive number of global image tokens, we propose the {\techfull} ({\tech}) strategy.
{\tech} explicitly guides object queries to attend to instance-relevant frames during training, providing structured supervision without extra inference overhead.
Extensive experiments demonstrate that {\methodname} achieves the state-of-the-art performance on ScanNetv2 and ScanNet200, outperforming both recent joint models and RGB-D-based approaches, while exhibiting strong generalization capabilities on ScanNet++.

\vspace{-1mm}

\end{abstract}

\section{Introduction}
\label{sec:intro}

\begin{figure}[t]
    \centering
        \includegraphics[width=0.98\linewidth]{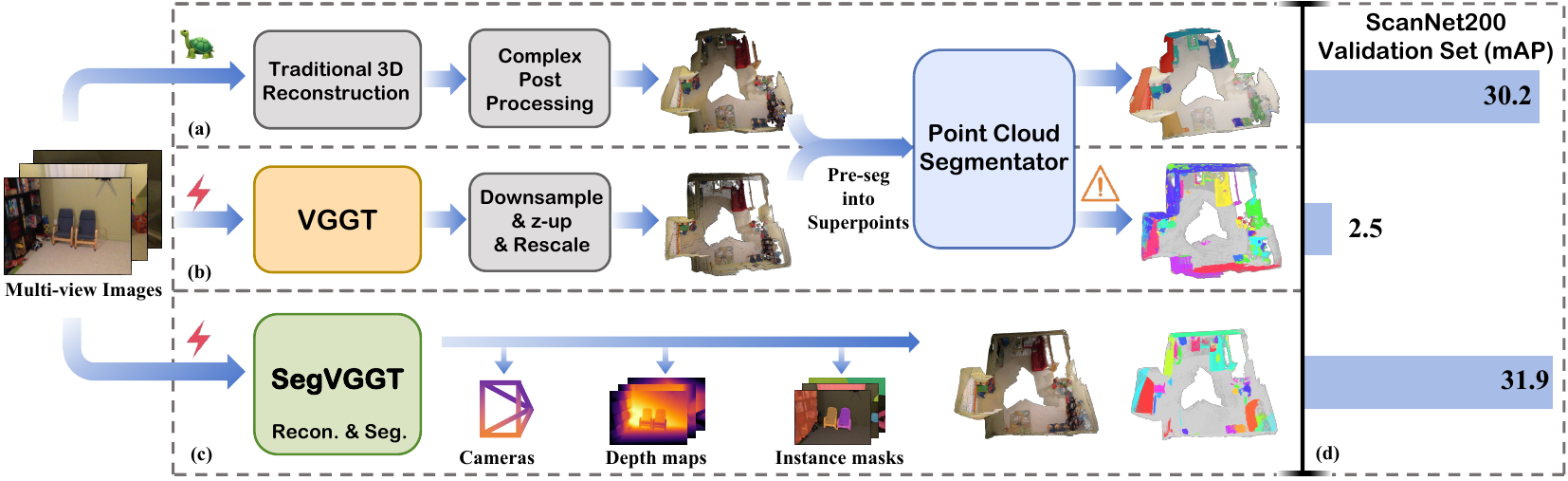}
    \caption{
        \textbf{Comparison of different 3D instance segmentation pipelines.}
        (a) Segmentation methods that rely on high-quality point clouds require complex and time-consuming processing, limiting their applicability.
        (b) Feed-forward 3D reconstruction improves efficiency, but leads to severely degraded segmentation performance.
        (c) Our proposed {\methodname} performs 3D reconstruction and instance segmentation simultaneously in a single forward pass, achieving superior segmentation quality.
        (d) 3D instance segmentation performance on the ScanNet200 validation set.
    } 
    \label{fig.teaser}
\end{figure}

Understanding 3D scenes at the instance level is fundamental for embodied perception~\cite{gu2024conceptgraphs}, robotics~\cite{zhuang2023instance}, AR/VR~\cite{park2020deep, jiang2025anysplat}, and autonomous navigation~\cite{xie2021unseen}.
Among various scene understanding tasks, 3D instance segmentation plays a central role, as it provides object-level geometric representations that support reasoning and interaction.
Current state-of-the-art 3D instance segmentation methods~\cite{qu2025segdino3d, li2025ovseg3r, lu2025relation3d, oneformer3d, ODIN, Mask3D, maft} predominantly rely on carefully reconstructed and post-processed point clouds or high-quality posed RGB-D scans as inputs, as shown in Fig.~\ref{fig.teaser} (a).
However, their overall pipeline typically involves multiple stages, including multi-view data acquisition, camera registration, fusion and reconstruction, and point cloud denoising before segmentation can be performed.
These fragmented workflows are usually time-consuming and considerably increase the system complexity, and their applicability is further limited when depth sensors are unavailable or reconstruction quality is imperfect.

Recent advances in 3D reconstruction have shifted the paradigm from traditional multi-stage pipelines~\cite{mur2017orb, schoenberger2016mvs, schoenberger2016sfm} to one-stage feed-forward transformer-based models~\cite{dust3r, leroy2024grounding, vggt, wang2025pi, keetha2025mapanything}, significantly improving reconstruction efficiency.
These approaches eliminate the need for iterative optimization and expensive post-processing, making large-scale 3D reconstruction practical.
Nevertheless, they primarily focus on geometry estimation and remain disconnected from high-level semantic understanding.
Consequently, 3D scene reconstruction and 3D semantic understanding are still treated as two relatively independent problems.

A straightforward strategy is to decompose the pipeline into two stages, where a feed-forward model first performs 3D reconstruction, followed by a point cloud-based segmentation model, as illustrated in Fig.~\ref{fig.teaser} (b).
However, such an approach normally suffers from two major limitations.
First, segmentation performance is highly sensitive to reconstruction quality.
Feed-forward reconstructions without heavy post-processing often contain noise and misalignment, causing point cloud segmentation methods to degrade significantly.
Second, decoupling reconstruction and segmentation limits the potential of leveraging geometric reasoning to enhance instance-level understanding, leading to suboptimal performance.

Although several recent works~\cite{zust2025panst3r, li2025iggt, koch2025unified} have explored joint modeling of 3D reconstruction and segmentation, existing approaches still exhibit some limitations.
Some of them retain a two-stage paradigm, treating the geometric foundation model~\cite{cabon2025must3r} as a frozen backbone and additionally introducing DINOv2~\cite{DINOv2} and MaskFormer~\cite{mask2former} for segmentation, which leads to suboptimal joint learning.
Others train an instance feature head solely with contrastive learning, which requires costly clustering during inference, cannot directly predict semantic categories, and often produces inaccurate segmentation boundaries.

In this work, we propose {\methodname}, a unified end-to-end framework which jointly performs feed-forward 3D reconstruction and 3D instance segmentation directly from multi-view RGB images, as shown in Fig.~\ref{fig.teaser} (c).
Our key insight is that instance understanding can be deeply integrated into the visual geometry grounded transformer.
To this end, we initialize a set of object queries and let them perform cross-attention with global image tokens after global self-attention in each transformer layer, enabling instance-level representations to progressively evolve together with multi-level geometric features throughout the network.
This tight coupling allows object queries to leverage both low-level spatial cues and high-level geometric abstractions.
We further employ a semantic feature head~\cite{ranftl2021vision} to produce cross-view consistent instance-level features, from which 3D instance masks are derived through dot products with the global object queries.

However, a typical scene may contain $10^4 \sim 10^5$ image tokens, and directly introducing object queries to interact with global image tokens without explicit guidance poses significant optimization challenges.
We empirically observe that the attention between queries and tokens becomes highly dispersed in this setting.
By contrast, transformer-based 3D instance segmentation models~\cite{Mask3D, maft, qu2025segdino3d} commonly employ mechanisms such as mask-attention or positional embeddings to guide object queries to attend to specific regions.
To address this issue within our unified 3D reconstruction and segmentation framework, we propose a {\techfull} ({\tech}) strategy.
Specifically, we introduce an additional supervision and incorporate a corresponding cost into the Hungarian matching process, encouraging object queries to assign higher attention weights to frames that contain the associated instance while interacting with global image tokens.
This mechanism provides a structured supervision for cross-frame attention without incurring additional inference overhead.

Extensive experiments validate the effectiveness of {\methodname}.
On the ScanNetv2~\cite{scannet} and ScanNet200~\cite{scannet200} benchmarks, our method achieves the state-of-the-art segmentation performance compared with both joint 3D reconstruction and segmentation models and RGB-D scan-based methods, demonstrating its superiority.
Moreover, without any additional training, we evaluate its performance on ScanNet++~\cite{yeshwanth2023scannet++}. 
Under the same setting, our method even surpasses approaches trained with ScanNet++, indicating strong generalization capability.

In summary, our contributions are threefold:
1) We present {\methodname}, a unified framework that simultaneously performs feed-forward 3D reconstruction and 3D instance segmentation directly from multi-view RGB images.
2) We integrate instance identification into the visual geometry grounded transformer through object queries and introduce the {\techfull} ({\tech}) strategy to efficiently solve the attention dispersion problem over massive image tokens without incurring inference overhead.
3) We achieve the state-of-the-art performance on ScanNetv2 and ScanNet200, surpassing recent joint models and even methods relying on posed RGB-D scans. Moreover, {\methodname} demonstrates strong generalization capabilities on ScanNet++.

\section{Related Works}
\label{sec:rel_works}

\noindent\textbf{3D Instance Segmentation on Point Clouds.}
3D instance segmentation aims to assign both semantic labels and distinct instance identities to 3D geometry. 
Early works either adopt a proposal-based paradigm that first generates 3D bounding box proposals and subsequently predicts instance masks~\cite{yang2019learning, hou20193d, yi2019gspn, engelmann20203d, td3d}, or follow a grouping-based strategy that clusters points into instances based on learned similarity without explicit proposal generation~\cite{liang2021instance, chen2021hierarchical, vu2022softgroup, jiang2020pointgroup, wang2019associatively, jiang2020end, zhang2021point}.
Recently, transformer-based methods~\cite{Mask3D, spformer, al20233d, competitorformer, maft, oneformer3d, queryformer, lu2025relation3d, qu2025segdino3d, li2025ovseg3r, SAI3D, open3dis} have become dominant by directly predicting instance masks with object queries.
These methods primarily operate on high-quality point clouds, which often require complex post-processing.
ODIN~\cite{ODIN} instead takes posed RGB-D scans as input, yet still assumes access to reconstructed 3D geometry.
Consequently, when only RGB images are available, these approaches are not directly applicable.

\noindent\textbf{Feed-forward 3D Reconstruction.}
Traditional 3D reconstruction frameworks typically follow a multi-stage pipeline involving feature matching, Structure-from-Motion (SfM)~\cite{schoenberger2016sfm, hartley2003multiple}, and Multi-View Stereo (MVS)~\cite{schoenberger2016mvs}. 
While being robust, these systems involve complex heuristic tuning and intensive computation, often requiring separate optimization phases to resolve geometric inconsistencies.
Recent advances, such as DUSt3R~\cite{dust3r} and its variants~\cite{leroy2024grounding, jang2025pow3r}, have shifted to a feed-forward paradigm that directly regresses dense point maps from image pairs.
VGGT~\cite{vggt} and its successors~\cite{wang2025pi, keetha2025mapanything} further propose architectures capable of processing hundreds of frames in a single forward pass to predict multiple geometric attributes, substantially improving efficiency and scalability.
Despite these advances, existing works focus primarily on geometric estimation and do not explicitly consider semantic or instance-level scene understanding.

\noindent\textbf{Joint 3D Reconstruction and Segmentation.}
Recently, there has been growing interest in bridging reconstruction and understanding in a unified model.
LSM~\cite{fan2024large} and Uni3R~\cite{sun2025uni3r} extend geometric foundation models~\cite{dust3r, vggt} with 3D Gaussian Splatting heads to produce semantic features, but they lack instance-level awareness.
SIU3R~\cite{xu2025siu3r} further incorporates instance perception, but supports only two-view inputs.
PanSt3R~\cite{zust2025panst3r} follows a two-stage paradigm, using a frozen MUSt3R~\cite{cabon2025must3r} model for scene reconstruction and treating it as a backbone combined with DINOv2~\cite{DINOv2} to extract features, which are then fed into a separate Mask2Former~\cite{mask2former} for panoptic segmentation, leading to suboptimal joint learning.
IGGT~\cite{li2025iggt} and UNITE~\cite{koch2025unified} perform segmentation by training an instance-level feature head with contrastive learning. However, they require computationally expensive clustering~\cite{mcinnes2017accelerated} during inference, which may yield inaccurate segmentation boundaries, and rely on additional VLMs~\cite{clip, li2022language, ghiasi2022scaling} to predict semantic categories.
MVGGT~\cite{wu2026mvggt} introduces the task of multi-view 3D referring expression segmentation along with a baseline model, aiming to reconstruct scenes from sparse views and segment the referred object.

\section{Method}
\label{sec:method}

\begin{figure}[t]
    \centering
        \includegraphics[width=0.98\linewidth]{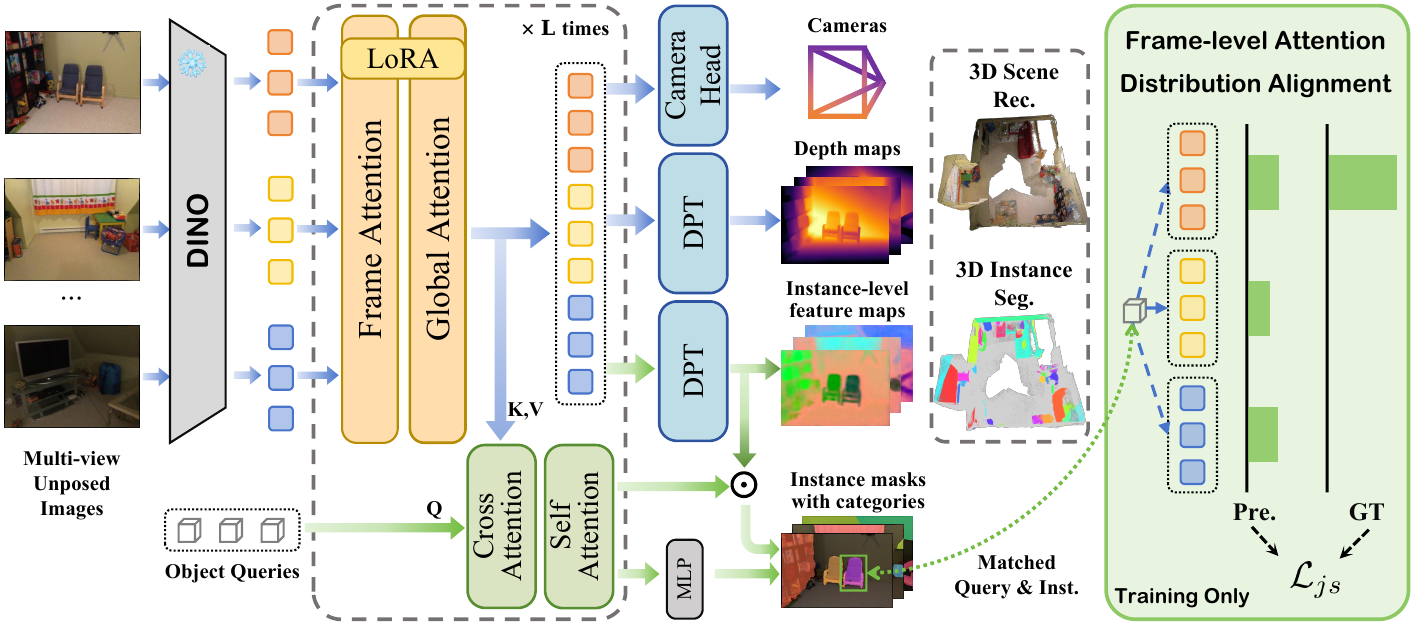}
    \caption{
        \textbf{Overall architecture of {\methodname}.}
        {\methodname} takes multi-view unposed RGB images as input and outputs the camera parameters, depth maps, and instance masks in a single forward pass.
        In the visual geometry grounded transformer, image tokens and object queries are progressively refined to capture both multi-view geometry and instance semantics.
        The green arrows indicate the components related to the 3D instance segmentation.
        During training, the {\techfull} ({\tech}) module regularizes the cross-attention weights via Jensen-Shannon Divergence loss $\mathcal{L}_{js}$, effectively mitigating the attention dispersion problem.
    } 
    \label{fig.method}
\end{figure}

\subsection{Overall Architecture}
\label{sec:overall}

As illustrated in Fig.~\ref{fig.method}, {\methodname} takes $N$ unposed RGB images $(I_i)_{i=1}^{N}$ as input, where $I_i \in \mathbb{R}^{3 \times H \times W}$ and $H \times W$ denotes the resolution of images.
Alongside the images, we also initialize a set of $O$ learnable object queries $(\mathbf{q}_j)^{O}_{j=1}$, where $\mathbf{q}_j \in \mathbb{R}^{d}$ and $d$ denotes the feature dimension.
Our unified transformer maps the input images to the corresponding camera parameters, depth maps, and instance-level feature maps in a single forward pass, while simultaneously updating the object queries. This process can be formulated as:
\begin{equation}
    f_{\theta}\Big( (I_i)_{i=1}^N, (\mathbf{q}_j)^{O}_{j=1} \Big) = \Big( (\mathbf{g}_i, D_i, F_i)_{i=1}^N, (\hat{\mathbf{q}}_j)^{O}_{j=1} \Big) ,
    \label{eq:transformer}
\end{equation}
Among the model outputs, $\mathbf{g}_i \in \mathbb{R}^{9}$ denotes the camera intrinsics and extrinsics, and $D_i \in \mathbb{R}^{H \times W}$ is the depth map with the same resolution as the input image.
$F_i \in \mathbb{R}^{d \times h \times w}$ represents the instance-level feature map, whose spatial resolution is half of the input image, i.e., $(h, w) = (\frac{H}{2}, \frac{W}{2})$, to reduce memory consumption.
$(\hat{\mathbf{q}}_j)^{O}_{j=1}$ denotes the optimized object queries after aggregating multi-level geometric features in the transformer.
From these outputs, {\methodname} can seamlessly derive instance segmentation masks with categories and obtain the final 3D reconstruction and 3D instance segmentation through simple unprojection.

\subsection{Geometry Grounded Transformer with Instance Reasoning}
\label{sec:aggregator}

To deeply integrate instance reasoning into the visual geometry grounded transformer, we extend this architecture by introducing a set of learnable object queries $(\mathbf{q}_j)^{O}_{j=1}$.
These queries are progressively refined across all the $L$ layers in the transformer, enabling them to leverage both low-level spatial cues and high-level geometric abstractions to enhance instance awareness.
Within the transformer, image tokens and object queries are optimized jointly.
The image tokens primarily capture multi-view geometric and semantic information and are used to regress dense outputs, while the object queries focus on instance-level understanding and reasoning.

\noindent\textbf{Update of Image Tokens.} 
Each input image $I_i$ is first processed by a frozen DINO~\cite{DINOv2} backbone to extract dense features, producing $K$ image tokens denoted as a matrix $T_i \in \mathbb{R}^{K \times d}$.
To construct a global representation of the 3D scene, the image tokens from all $N$ views are concatenated along the sequence dimension to form an aggregated multi-view token matrix $T \in \mathbb{R}^{(NK) \times d}$, where $T =[T_1; T_2; \dots; T_N]$\footnote{In practice, each image is also augmented with a camera token and four register tokens, which are omitted here and in Fig.~\ref{fig.method} for clarity.}.
These tokens are then fed into the unified transformer, where features are fused and updated through alternating frame attention and global attention.
We denote the multi-view image tokens after optimization by the $l$-th ($1 \leq l \leq L$) global attention layer as $T^{(l)}$, where $T^{(0)}$ corresponds to the original concatenated image tokens extracted by DINO.

\noindent\textbf{Update of Object Queries.} 
Different from prior works that focus solely on geometric estimation, {\methodname} aims to construct a transformer architecture that jointly captures geometric and instance-level information.
To this end, we initialize $O$ learnable object queries $(\mathbf{q}_j)^{O}_{j=1}$ and update them throughout the transformer layers.
Specifically, after the image tokens $T^{(l)}$ undergo global attention for cross-view feature fusion, we insert a cross-attention module that enables the object queries to extract information from them.
With the $j$-th object query acting as the query and the unified image tokens acting as the keys and values, this cross-attention process at the $l$-th layer can be formulated as:
\begin{equation}
    A_j^{(l)} = \operatorname{Softmax}\left( \mathbf{q}_j^{(l-1)} (T^{(l)})^\top / \sqrt{d} \right), 
    \quad
    \mathbf{q}_j^{\prime (l)} = \mathbf{q}_j^{(l-1)} + A_j^{(l)} T^{(l)},
    \label{eq:cross_attn}
\end{equation}
where $A_j^{(l)} \in \mathbb{R}^{1 \times (NK)}$ represents the global cross-attention weight map of the $j$-th query over all multi-view image tokens, and $\mathbf{q}_j^{\prime (l)} \in \mathbb{R}^{1 \times d}$ denotes the intermediate query updated by the cross-attention.
Subsequently, a self-attention layer among the queries is applied to further enhance their feature representations, yielding the final output $\mathbf{q}_j^{(l)}$ for the current layer.

\subsection{Synchronous 3D Decoding and Training Objective}
\label{sec:decoding}

Following the $L$ layers of the geometry grounded transformer, the enriched image tokens $T^{(L)}$ and object queries $\mathbf{q}_j^{(L)}$ are fed into the decoding heads to yield the final 3D scene reconstruction and instance segmentation.

\noindent\textbf{Geometry Predictions.} 
To reconstruct the multi-view geometry, we follow VGGT~\cite{vggt} by employing a Camera Head to regress the camera parameters $(\mathbf{g}_i)^{N}_{i=1}$, and a DPT~\cite{ranftl2021vision} head to predict the dense depth maps $(D_i)^{N}_{i=1}$.

\noindent\textbf{Instance Predictions.}
For instance-level understanding, a separate semantic DPT head is introduced to decode the image tokens into cross-view consistent instance-level feature maps $F_i \in \mathbb{R}^{d \times h \times w}$.
Meanwhile, each optimized query $\hat{\mathbf{q}}_j = \mathbf{q}_j^{(L)}$ predicts semantic logits $\mathbf{c}_j \in \mathbb{R}^{C+1}$ via an MLP classification head, where $C$ is the number of instance categories.
To generate the multi-view 2D masks, we compute the dot product between each object query $\hat{\mathbf{q}}_j$ and the dense feature maps $(F_i)_{i=1}^N$. 
The probability mask $M_{j, i} \in [0, 1]^{h \times w}$ for the $j$-th query on the $i$-th view is formulated as:
\begin{equation}
    M_{j, i} = \sigma\left(\hat{\mathbf{q}}_j^\top F_i\right),
\end{equation}
where $\sigma$ denotes the sigmoid function. During inference, a pre-defined threshold $\tau_{m}$ is applied to binarize $M_{j, i}$.
With the synchronously predicted camera parameters $\mathbf{g}_i$ and depth maps $D_i$, these binarized 2D masks can be directly unprojected into a unified 3D coordinate space, elegantly yielding the final 3D instance segmentation without relying on any complex point cloud post-processing.

\noindent\textbf{Training Objective.}
The overall training objective $\mathcal{L}_{total}$ is composed of the geometric loss, the instance loss, and our proposed attention alignment loss:
\begin{equation}
    \mathcal{L}_{total} = \mathcal{L}_{geo} + \mathcal{L}_{inst} + \lambda_{js} \mathcal{L}_{js}.
\end{equation}

For 3D scene reconstruction, $\mathcal{L}_{geo} = \lambda_{camera} \mathcal{L}_{camera} + \lambda_{depth} \mathcal{L}_{depth}$ is used to supervise the predicted camera parameters $(\mathbf{g})^{N}_{i=1}$ and depth maps $(D_i)^{N}_{i=1}$.
Both $\mathcal{L}_{camera}$ and $\mathcal{L}_{depth}$ directly follow the design in VGGT~\cite{vggt}, and their detailed formulations are omitted here for brevity.
Notably, we adopt a pre-trained VGGT model with frozen parameters as a teacher model to provide geometric supervision, which aims to mitigate overfitting of geometry predictions to noise and biases in the training data.

For 3D instance segmentation, we follow DETR~\cite{detr} and employ the Hungarian algorithm~\cite{kuhn1955hungarian} to match object queries with ground-truth instances, enabling end-to-end training with classification and mask supervision.
Interestingly, we observe that by simply flattening the predicted multi-view masks $M_j \in[0, 1]^{N \times H \times W}$ of each query into a 1D sequence $m_j \in [0, 1]^{NHW}$, the representation becomes mathematically equivalent to a 3D point cloud mask. 
By applying the identical operation to the multi-view consistent 2D ground-truth masks, we can seamlessly inherit the matching cost and loss formulations from point cloud segmentation methods~\cite{spformer, oneformer3d}. 
Specifically, the pair-wise matching cost $C_{j, k}$ between the $j$-th prediction and the $k$-th ground truth is defined as:
\begin{equation}
    C_{j, k} = -\lambda_{cls} \, c_{j, c_k} + \lambda_{mask} \left( \text{BCE}(m_j, m_k^{gt}) + \text{Dice}(m_j, m_k^{gt}) \right) + \lambda_{js} \, C_{j, k}^{js},
\end{equation}
where $c_{j, c_k}$ denotes the predicted probability of the $j$-th query belonging to the target class $c_k$, and $m_k^{gt} \in \{0, 1\}^{NHW}$ is the flattened ground-truth mask.
Here, $\text{BCE}(\cdot, \cdot)$ and $\text{Dice}(\cdot, \cdot)$ represent the binary cross-entropy and dice loss with Laplace smoothing~\cite{milletari2016v}.
Crucially, $C_{j, k}^{js}$ represents our proposed frame-level attention alignment cost based on the Jensen-Shannon divergence, which explicitly encourages queries to match with instances that geometrically align with their attention priors (detailed in Sec.~\ref{sec:tech}). 
Once the optimal assignment is established, the instance loss $\mathcal{L}_{inst}$ is computed over the matched pairs as a weighted sum of the classification loss and the mask segmentation losses:
\begin{equation}
    \mathcal{L}_{inst} = \lambda_{cls} \mathcal{L}_{cls} + \lambda_{mask} \left( \mathcal{L}_{bce} + \mathcal{L}_{dice} \right).
\end{equation}

Finally, $\mathcal{L}_{js}$ is our proposed frame-level attention alignment loss, which acts as a structural regularization during training and will be detailed in Sec.~\ref{sec:tech}.

\subsection{{\techfull}}
\label{sec:tech}

A typical 3D scene inherently involves a massive number of aggregated multi-view image tokens (e.g., $NK \approx 10^4 \sim 10^5$) in our architecture. 
In such a high-dimensional space, the cross-attention weights of the object queries tend to become highly dispersed across the entire sequence. 
However, due to occlusion and limited fields of view, a specific 3D instance is typically visible in only a sparse subset of the $N$ views. 
This mismatch further amplifies attention dispersion.
Without explicit constraints, the unguided object queries normally struggle to localize the correct views and are prone to absorbing irrelevant background noise, leading to optimization challenges and suboptimal performance.

To bridge this gap, we propose the {\techfull} ({\tech}) module to provide an explicit global-to-local guidance.
Our core insight is to directly align the queries' attention distribution with the actual visibility of the target instances across the multi-view frames.
Instead of solely applying this constraint at the final output, we employ deep supervision across the intermediate transformer layers to progressively refine the queries' geometric awareness.
Specifically, at the $l$-th transformer layer, we extract the cross-attention weight map $A_j^{(l)} \in \mathbb{R}^{1 \times (NK)}$ of the $j$-th query obtained from Equation~\eqref{eq:cross_attn}.
Since $A_j^{(l)}$ is a normalized probability distribution over all tokens, we can naturally obtain the predicted frame-level attention distribution $\hat{\mathbf{p}}_j^{(l)} \in [0,1]^{N}$ by marginalizing the weights over the spatial dimension of each view. 
For the $i$-th view $I_i$, its aggregated attention score is calculated as:
\begin{equation}
    \hat{p}_{j, i}^{(l)} = \sum_{t \in \text{view}_i} A_{j, t}^{(l)}.
\end{equation}

Meanwhile, for a ground-truth instance $k$, we deterministically construct its target visibility distribution $\mathbf{p}_k^{gt} \in [0,1]^{N}$.
Specifically, the target probability $p_{k, i}^{gt}$ for the $i$-th view $I_i$ is defined proportionally to the number of pixels the instance occupies in that frame relative to its total pixel count across all $N$ views. 
This area-proportional distribution ensures that frames containing a larger visible portion of the target instance receive correspondingly higher attention priors, while unobserved frames are naturally assigned a probability of zero.
To measure the alignment between the predicted attention focus and the ground-truth visibility, we utilize the Jensen-Shannon (JS) divergence.
Unlike the Kullback-Leibler (KL) divergence, the JS divergence is symmetric, inherently bounded, and numerically stable for distributions containing zero probabilities. 
The proposed {\tech} module plays a dual role in our end-to-end framework.

\noindent\textbf{Prior-guided Matching.}
During the bipartite matching phase, we incorporate the JS divergence across all $L$ transformer layers as a structural prior. 
To ensure numerical stability across different sequence lengths and network depths, the matching cost $C_{j, k}^{js}$ is normalized by the total number of frames $N$ and layers $L$:
\begin{equation}
    C_{j, k}^{js} = \frac{1}{L \cdot N} \sum_{l=1}^{L} \text{JS}(\mathbf{p}_k^{gt} \parallel \hat{\mathbf{p}}_j^{(l)}).
\end{equation}
This explicitly encourages queries to be assigned to instances that intrinsically align with their geometric attention focus throughout the entire progressive refinement process, easing the optimization difficulty.

\noindent\textbf{Attention Regularization.}
After the optimal matching $\mathcal{M}$ is established, we apply the alignment loss $\mathcal{L}_{js}$ to penalize distribution inconsistency between matched pairs. 
Consistent with the matching cost, the overall alignment loss is averaged over all layers and frames:
\begin{equation}
    \mathcal{L}_{js} = \frac{1}{|\mathcal{M}| \cdot L \cdot N} \sum_{(j, k) \in \mathcal{M}} \sum_{l=1}^{L} \text{JS}(\mathbf{p}_k^{gt} \parallel \hat{\mathbf{p}}_j^{(l)}).
\end{equation}

By imposing this structured deep guidance both in the assignment and optimization stages, {\methodname} enforces the object queries to firmly anchor on instance-relevant frames during progressive cross-view feature fusion. 
Notably, the computational overhead introduced by this module is negligible during training.
Rather than requiring additional parameterized layers, it directly reuses the cross-attention weights inherently computed by the transformer.
Furthermore, operating at the frame level significantly reduces the dimensionality of calculation, making it highly efficient. 
Crucially, as this module acts strictly as a regularization mechanism during training, it is completely removed during inference and introduces no additional computational cost or network parameters.

\section{Experiments}
\label{sec:exp}

\subsection{Experiment Setup}
\label{sec:exp_setup}

\noindent\textbf{Datasets.} 
We evaluate {\methodname} on three widely adopted indoor datasets: ScanNetv2~\cite{scannet}, ScanNet200~\cite{scannet200}, and ScanNet++~\cite{yeshwanth2023scannet++}.
ScanNetv2 comprises 1,613 3D indoor scenes with dense semantic and instance annotations, covering 20 semantic classes and 18 instance categories.
Following the official split, it contains 1,201 training scenes, 312 validation scenes, and 100 test scenes.
ScanNet200 shares the same underlying scenes and data splits as ScanNetv2, but significantly expands the label space from 20 to 200 fine-grained semantic categories, among which 198 are instance categories.
It exhibits a severe long-tail distribution, presenting a much greater challenge for instance-level understanding.
ScanNet++ is a recently introduced high-fidelity dataset comprising 856, 50, and 50 scenes for training, validation, and testing, respectively, with 84 categories for evaluating 3D instance segmentation.
All three datasets provide point clouds, RGB-D images, and corresponding camera intrinsics and extrinsics.
In these experiments, we train and evaluate on ScanNetv2 and ScanNet200, respectively, while ScanNet++ is used exclusively to assess the generalization capability of our method without any fine-tuning on it.

\noindent\textbf{Implementation Details.}
We build {\methodname} upon the VGGT~\cite{vggt} architecture with $L=24$ and initialize it with pre-trained weights. 
During training, the DINO backbone is kept frozen.
To optimize computational efficiency, we employ LoRA~\cite{hu2022lora} to fine-tune the frame attention and global attention modules, while fully training the newly introduced parameters.
For each training iteration, we randomly sample 2-24 frames per scene as input. 
For geometric supervision, we utilize a frozen, pre-trained VGGT as a teacher model to distill robust depth maps and camera parameters.
The instance segmentation branch is supervised using multi-view consistent 2D masks provided by the datasets, which are generated by projecting the 3D instance annotations onto the respective 2D frames. 
We train the models separately on ScanNetv2 and ScanNet200 using 8 NVIDIA A100 GPUs, taking approximately 2 days per dataset.
For more implementation details, please refer to the appendix.

\noindent\textbf{Evaluation Metrics.}
We follow the standard evaluation metric for 3D instance segmentation and report mean Average Precision (mAP), defined as the area under the precision-recall curve across multiple Intersection-over-Union (IoU) thresholds.
Concretely, mAP is averaged over IoU thresholds from 50\% to 95\% with a step of 5\%, providing a comprehensive measure of segmentation quality.
We also report mAP$_{50}$ and mAP$_{25}$ at fixed IoU thresholds of 50\% and 25\%, respectively, to evaluate performance under different localization tolerances.

\noindent\textbf{Evaluation Settings.}
To comprehensively evaluate {\methodname}, we compare it against state-of-the-art 3D instance segmentation methods under two primary input modalities:
1) Point Cloud-based Methods~\cite{qu2025segdino3d, oneformer3d, ODIN, Mask3D, maft, lu2025relation3d, queryformer}.
As the dominant paradigm, these approaches predominantly operate on point clouds sampled from benchmark meshes, constituting a privileged setting with direct access to accurate 3D geometry.
This avoids reconstruction artifacts and projection inconsistencies, while assuming pre-reconstructed 3D inputs that may not be readily available in practical deployments.
ODIN~\cite{ODIN} instead takes posed RGB-D scans provided by the benchmark as input.
Although less restrictive, it still relies on additional modalities beyond RGB images.
2) RGB Image-based Methods~\cite{zust2025panst3r, li2025iggt}.
Approaches in this category, including ours, tackle a more practical yet challenging setting by predicting 3D instances purely from 2D images.
Since ground-truth instance masks are defined on the official 3D meshes, predicted multi-view masks must be mapped to these benchmark point clouds for evaluation.
We utilize the ground-truth depth maps and camera poses during this mapping stage for fair comparison.
Nevertheless, this projection step inevitably introduces noise and misalignments, placing RGB-based approaches at an inherent disadvantage compared to point cloud-based baselines.

\begin{table}[t]
    \centering
    \caption{
        \textbf{Comparison of {\methodname} with point cloud-based methods.}
        Input modalities are denoted as follows: P (point cloud from the benchmark), I (RGB images), D (depth maps), and C (camera parameters).
        $^{\dagger}$ indicates that the method first performs 3D reconstruction from the input images with VGGT~\cite{vggt}, and then performs 3D instance segmentation on the reconstructed point cloud.
        }
    \label{tab:main_point_cloud}
    \begin{tabular}{l|c|ccc|ccc}
    \toprule

    &  & \multicolumn{3}{c|}{ScanNetv2} & \multicolumn{3}{c}{ScanNet200} \\
    Method & Input Modality & mAP & mAP$_{50}$ & mAP$_{25}$ & mAP & mAP$_{50}$ & mAP$_{25}$ \\

    \midrule
    Mask3D~\cite{Mask3D} & P & 55.2 & 73.7 & 85.3 & 27.4 & 37.0 & 42.3 \\
    QueryFormer~\cite{queryformer} & P & 56.5 & 74.2 & 83.3 & 28.1 & 37.1 & 43.4 \\
    MAFT~\cite{maft} & P & 59.9 & 76.5 & - & 29.2 & 38.2 & 43.3 \\
    SPFormer~\cite{spformer} & P & 56.3 & 73.9 & 82.9 & - & - & - \\
    SGIFormer-L~\cite{yao2024sgiformer} & P & 61.0 & 81.2 & \textbf{88.9} & 29.2 & 39.4 & 44.2 \\
    OneFormer3D~\cite{oneformer3d} & P & 59.3 & 78.1 & 86.4 & 30.2 & 40.9 & 44.6 \\
    Relation3D~\cite{lu2025relation3d} & P  & 62.5 & 80.2 & 87.0 & 31.6 & 41.2 & 45.6 \\
    OpenMask3D~\cite{openmask3d} & P \& I \& D \& C & - & - & - & 15.4 & 19.9 & 23.1 \\
    Open-YOLO 3D~\cite{openyolo3d} & P \& I \& D \& C & - & - & - & 24.7 & 31.7 & 36.2 \\
    SegDINO3D~\cite{qu2025segdino3d} & P \& I \& D \& C & \textbf{64.0} & \textbf{81.5} & \textbf{88.9} & \textbf{40.2} & \textbf{52.4} & \textbf{58.6} \\

    \midrule
    ODIN~\cite{ODIN} & I \& D \& C & 50.0 & 71.0 & 83.6 & 31.5 & 45.3 & 53.1 \\
    OneFormer3D$^{\dagger}$~\cite{oneformer3d} & I & 5.4 & 10.2 & 17.4 & 2.5 & 4.1 & 6.4 \\
    
    \textbf{Ours} & I & \textbf{50.4} & \textbf{71.7} & \textbf{87.0} & \textbf{31.9} & \textbf{45.7} & \textbf{53.7} \\

    \bottomrule
    \end{tabular}
\end{table}

\subsection{Comparison with Point Cloud-Based Methods}
\label{sec:pointcloud_based_comparison}

We first perform a class-aware comparison with standard point cloud-based 3D instance segmentation methods on the ScanNetv2 and ScanNet200 validation sets (Table~\ref{tab:main_point_cloud}).
We categorize baselines by input modalities.
While conventional methods directly operate on benchmark-provided point clouds and some incorporate additional modalities, ODIN~\cite{ODIN} avoids pre-reconstructed meshes but still relies on accurate sensor point clouds to provide geometric information for inference.
In contrast, {\methodname} is the only method relying exclusively on unposed 2D RGB images (uniformly sampled every 20 frames as input).
This imposes the most stringent evaluation challenges, including inevitable projection misalignments and entirely unobserved ground-truth regions.

\noindent\textbf{Methods without Benchmark Point Clouds.}
Among methods lacking direct access to the benchmark point clouds (bottom of Table~\ref{tab:main_point_cloud}), {\methodname} achieves state-of-the-art results across all metrics on both datasets, despite operating under the strictest input constraints.
To further validate the necessity of a unified architecture, we construct a naive two-stage pipeline that combines VGGT~\cite{vggt} with a widely used point cloud segmentation model, OneFormer3D~\cite{oneformer3d}.
We observe a drastic performance degradation, with mAP on ScanNet200 decreasing from 30.2 to 2.5.
This drop is partially caused by evaluation projection errors, but more fundamentally, it highlights that existing point cloud segmentation methods are overly sensitive to reconstruction noise and heavily rely on pristine geometry and cumbersome post-processing.

\noindent\textbf{Methods on Benchmark Point Clouds.}
When compared to methods directly segmenting the benchmark point clouds, {\methodname} exhibits a gap in the strict mAP metric on ScanNetv2, but surpasses most approaches under the looser mAP$_{25}$ threshold and achieves performance comparable to Relation3D~\cite{lu2025relation3d}.
Consistent with observations in ODIN~\cite{ODIN}, we attribute this gap under strict metrics to inevitable projection errors that lead to misalignments with the ground-truth point clouds during evaluation.
The strong mAP$_{25}$ performance indicates that our model retains robust instance identification capability despite the evaluation mismatch.
On the significantly more challenging ScanNet200 benchmark, our method showcases a compelling advantage in handling fine-grained categories, outperforming all pure point cloud-based methods. 
While SegDINO3D~\cite{qu2025segdino3d} achieves higher metrics, it relies on richer input modalities and additionally incorporates a powerful pre-trained 2D perception model~\cite{DINOX}.
In contrast, our RGB-only framework achieves robust performance with a much more elegant and practical deployment setup.

\subsection{Comparison with RGB Image-Based Methods}
\label{sec:image_based_comparison}

\begin{table}[t]
    \centering
    \caption{
        \textbf{Comparison of {\methodname} with RGB image-based methods.}
        }
    \label{tab:main_image}
    \begin{tabular}{l|ccc|ccc|ccc}
    \toprule

    & \multicolumn{3}{c|}{ScanNetv2} & \multicolumn{3}{c|}{ScanNet200} & \multicolumn{3}{c}{ScanNet++} \\
    Method & mAP & mAP$_{50}$ & mAP$_{25}$ & mAP & mAP$_{50}$ & mAP$_{25}$ & mAP & mAP$_{50}$ & mAP$_{25}$ \\

    \midrule

    PanSt3R~\cite{zust2025panst3r} & 26.9 & 52.6 & 63.9 & 20.8 & 41.4 & 59.0 & 9.2 & 22.4 & 36.2 \\
    IGGT~\cite{li2025iggt} & 29.2 & 43.0 & 55.1 & 28.2 & 42.4 & 55.3 & 11.5 & 21.3 & 36.0 \\
    \textbf{Ours} & \textbf{62.9} & \textbf{87.4} & \textbf{97.4} & \textbf{53.7} & \textbf{80.2} & \textbf{89.9} & \textbf{13.3} & \textbf{33.9} & \textbf{56.4} \\

    \bottomrule
    \end{tabular}
\end{table}

We further compare our method with recent RGB image-based approaches, PanSt3R~\cite{zust2025panst3r} and IGGT~\cite{li2025iggt}, which concurrently perform 3D reconstruction and instance segmentation. 
Because IGGT relies on a clustering algorithm~\cite{mcinnes2017accelerated} to generate instance masks and cannot directly predict semantic categories, we conduct evaluations under a class-agnostic setting to ensure a fair comparison.
Furthermore, due to the slow inference speed of the baseline methods, full-set evaluation is computationally expensive.
To address this, we randomly sample 10 scenes from the validation set of each benchmark and apply uniform downsampling to the input frames for all methods: a $40\times$ rate for ScanNetv2 and ScanNet200, and a $160\times$ rate for the denser ScanNet++ scenes.

As shown in Table~\ref{tab:main_image}, {\methodname} achieves state-of-the-art performance across all datasets.
Specifically, it delivers substantial improvements on the ScanNetv2 and ScanNet200 benchmarks, demonstrating the efficacy of our model in purely vision-based 3D scene understanding.
To further validate its robustness and generalization capability, we evaluate our model on the high-fidelity ScanNet++ dataset.
Notably, while the baseline methods are trained on massive datasets that explicitly include ScanNet++ training scenes, our model is trained solely on ScanNet200 and applied to ScanNet++ without any fine-tuning.
Despite this setting, {\methodname} consistently outperforms these baselines across all metrics.
Particularly in localization precision, our method outperforms IGGT by 20.4 on the mAP$_{25}$ metric, demonstrating strong cross-dataset generalization ability.
Overall, these results highlight the structural advantages of our query-based unified framework. 
By deeply coupling instance reasoning with geometric features, it effectively addresses the insufficient joint modeling present in 
PanSt3R, while elegantly bypassing the high-cost and heuristically-tuned post-clustering required by IGGT during inference.

\subsection{Ablation Studies and Analysis}
\label{sec:ablation}

\begin{figure}[t]
    \centering
        \includegraphics[width=0.98\linewidth]{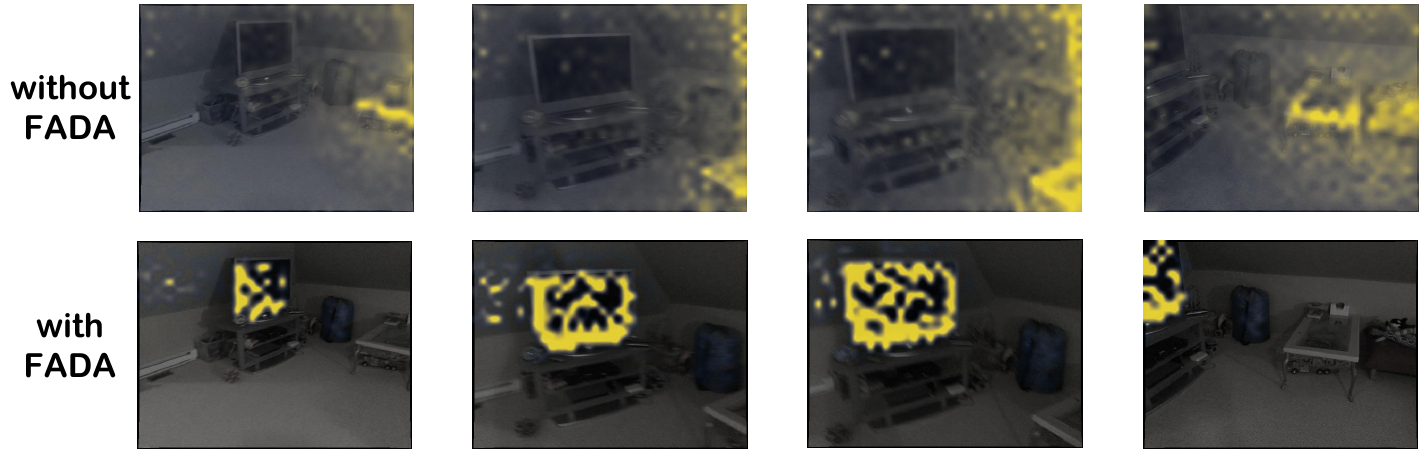}
    \caption{
        \textbf{Visualization of the {\tech} module's effect.}
        We visualize the similarity map between the query corresponding to the TV instance and the image tokens in the last cross-attention layer.
        Without the {\tech} module, the query attention is dispersed across the entire scene.
        After introducing {\tech}, the query focuses much more accurately on the TV, indicating that {\tech} improves the model's ability to attend to target instances and thereby enhances instance segmentation performance.
    } 
    \label{fig.tech}
\end{figure}

We conduct ablation studies on the ScanNet200 validation set to evaluate the effectiveness of different components in our method.

\noindent\textbf{Ablation on {\tech} Module.}
As described in Sec.~\ref{sec:tech}, we introduce the {\tech} module to mitigate the attention dispersion problem. 
To evaluate its effectiveness, we investigate its impact by decoupling its dual roles: attention regularization through the loss term and prior-guided matching through the cost term.
As shown in the first row of Table~\ref{tab:ab_tech}, without any explicit guidance, the baseline model only achieves 26.7 mAP, indicating that unguided queries struggle to accurately segment instances.
By introducing the JS divergence as a regularization loss during training, the performance substantially improves by +4.4 mAP, as shown in row 2.
This significant gain validates that structurally aligning the attention distribution with target visibility effectively mitigates the attention dispersion issue.
Furthermore, incorporating the JS divergence into the Hungarian matching cost yields the best performance of 31.9 mAP, as shown in row 3.
This confirms that leveraging geometric attention priors to guide bipartite matching successfully reduces optimization difficulty, thereby further enhancing the overall accuracy and localization precision.

\begin{table}[tbh]
\parbox{.5\linewidth}{
    \centering
    \caption{
        \textbf{Ablation on the {\tech} module.}
        We decouple the {\tech} module into loss supervision and cost function.
    }
    \label{tab:ab_tech}
    \begin{tabular}{c|c|c|ccc}
    \toprule
    Row & Loss & Cost & mAP & mAP$_{50}$ & mAP$_{25}$ \\
    \midrule
    1 & \ding{55}   & \ding{55}     & 26.7 & 39.2 & 48.6 \\
    2 & \ding{51}   & \ding{55}     & 31.1 & \textbf{45.7} & 52.3 \\
    3 & \ding{51}   & \ding{51}     & \textbf{31.9} & \textbf{45.7} & \textbf{53.7} \\
    \bottomrule
    \end{tabular}
}
\hfill
\parbox{.46\linewidth}{
    \centering
    \caption{
        \textbf{Ablation on where to insert query interaction}
        }
    \label{tab:ab_layers}
    \begin{tabular}{l|ccc}
    \toprule
    Inserted Layers & mAP & mAP$_{50}$ & mAP$_{25}$ \\
    \midrule
    Early 12   & 23.8 & 36.9 & 45.6 \\
    Late 12    & 30.5 & 45.1 & 53.0 \\
    Interleaved    & 30.4 & 44.5 & 51.8 \\
    All 24    & \textbf{31.9} & \textbf{45.7} & \textbf{53.7} \\
    \bottomrule
    \end{tabular}
}
\end{table}

In addition, we further analyze the effect of the {\tech} module through visualization.
We train two models, one with the {\tech} module and one without it, and identify the query with the highest score for the TV instance in the same scene.
As shown in Fig.~\ref{fig.tech}, we visualize the similarity maps between the selected query and the global image tokens in the last cross-attention layer, and present four views.
The results show that without the {\tech} module, the query attention is highly dispersed and often focuses on incorrect regions.
In contrast, after introducing the {\tech} module, the query attention accurately concentrates on the TV instance.
This observation further confirms that the {\tech} module effectively mitigates attention dispersion and guides the queries to focus on the target instance, thereby improving instance segmentation performance.

\noindent\textbf{Ablation on Multi-level Geometric Features.}
In {\methodname}, object queries are progressively updated through cross-attention with image tokens.
To validate the necessity of this deep integration, we ablate the insertion locations of these query interactions within the 24-layer transformer. 
As illustrated in Table~\ref{tab:ab_layers}, constraining interactions only to the early 12 layers yields a poor mAP of 23.8, as early tokens primarily capture local 2D appearance cues without an established multi-view geometric consensus. 
Conversely, interacting solely with the late 12 layers improves the mAP to 30.5, highlighting the critical role of global 3D geometric information for instance reasoning. 
Similarly, interleaving 12 interaction layers throughout the network yields 30.4 mAP.
Ultimately, inserting the interaction modules across all 24 layers achieves the optimal performance. 
This clearly demonstrates that object queries benefit most when they continuously co-evolve with the image tokens, seamlessly absorbing multi-level features ranging from low-level spatial details to high-level geometric structures.

\section{Conclusion}

In this work, we presented {\methodname}, a unified feed-forward framework that simultaneously performs 3D scene reconstruction and 3D instance segmentation directly from unposed multi-view RGB images in a single forward pass.
By deeply integrating object queries into a visual geometry grounded transformer, instance-level representations progressively co-evolve with multi-view geometric features, enabling tight coupling between geometry estimation and instance reasoning.
To address the severe attention dispersion arising from massive global image tokens, we propose the {\techfull} ({\tech}) module, which serves a dual role as a structured bipartite matching prior and an attention regularization loss, explicitly anchoring queries to instance-relevant frames without any extra inference overhead.
Extensive experiments demonstrate that {\methodname} achieves state-of-the-art performance among RGB image-based methods, surpasses prominent point cloud-native baselines on the challenging ScanNet200 benchmark, and exhibits strong generalization on ScanNet++.

\bibliographystyle{splncs04}
\bibliography{main}

\appendix 
\clearpage
\section{Appendix}

The appendix is organized as follows.
Sec.~\ref{sec:efficiency} presents a detailed efficiency analysis of {\methodname}, including scalability evaluations with varying input lengths and runtime comparisons against recent joint reconstruction and segmentation methods.
Sec.~\ref{sec:more_ablation} provides additional ablation studies examining the contributions of joint optimization and unified architecture design.
Sec.~\ref{sec:more_vis_tech} provides more complete visualizations illustrating the effect of the {\tech} module on cross-attention maps across transformer layers.
Sec.~\ref{sec:more_details} supplements the implementation details in the main paper with full specifications of model configuration, data augmentation, optimization strategy, and loss function weights.
Sec.~\ref{sec:more_eval_details} elaborates on the evaluation protocols, including the 2D-to-3D mapping strategy and the setup of the two-stage and RGB image-based baselines.
Sec.~\ref{sec:geometric_performance} evaluates the geometric reconstruction capabilities of our method, demonstrating that the joint training paradigm maintains robust 3D geometry without negative transfer.
Finally, Sec.~\ref{sec:limitations} discusses the current limitations of {\methodname} and outlines promising directions for future work, and Sec.~\ref{sec:qualitative} presents qualitative results on joint 3D reconstruction and instance segmentation.

\subsection{Efficiency Analysis}
\label{sec:efficiency}
In this subsection, we analyze the efficiency of {\methodname} and provide a detailed comparison with existing methods that jointly perform 3D reconstruction and instance segmentation.

\noindent\textbf{Evaluation Protocol.} 
All efficiency evaluations are conducted on a single NVIDIA A100 GPU. 
For a fair assessment, we define the total runtime as the duration from receiving a sequence of multi-view RGB images to producing all information required for unified 3D scene understanding.
This includes geometry reconstruction outputs such as depth maps and camera poses, or point maps depending on the method, together with the final 3D instance masks.
We measure the inference time and post-processing time separately.
To ensure stable profiling, we execute 1 warmup run followed by 3 repeated runs and report the average time. 
For {\methodname} and IGGT~\cite{li2025iggt}, we set the chunk size of the DPT head to 8 to balance memory usage and inference speed.
The input image resolutions align with the official settings of each method: $518 \times 392$ for our method, $504 \times 336$ for IGGT, and $512 \times 512$ for PanSt3R~\cite{zust2025panst3r}.

\noindent\textbf{Scalability Analysis of {\methodname}.} 
We first evaluate the scalability of our approach by varying the number of input frames from 10 to 200.
As shown in Table~\ref{tab:efficiency_ours}, the inference time is decomposed into the transformer aggregator and the decoding heads.
The post-processing stage, which converts the raw model outputs into the final 3D reconstruction and instance segmentation results, is highly efficient in our framework.
In practice, it only requires the object queries to predict semantic categories and compute dot products with the instance-level feature maps to produce masks, together with lightweight score calculations.
Compared with conventional pipelines that rely on extensive post-processing to obtain refined 3D geometry before performing instance segmentation, our method avoids these costly stages and instead performs 3D reconstruction and instance segmentation jointly in a single feed-forward process, resulting in low computational overhead that scales well with the input length.
Consequently, processing a standard sequence of 100 frames takes only 15.9 seconds in total, indicating strong system-level efficiency.

\begin{table}[tbh]
    \centering
    \caption{
        \textbf{Scalability and efficiency analysis of {\methodname}.}
        We report the runtime breakdown, including aggregator, decoding heads, post-processing, total runtime, and peak reserved memory under different numbers of input frames.
    }
    \label{tab:efficiency_ours}
    \begin{tabular}{c|ccc|c|c|c}
        \toprule
        \multirow{2}{*}{Input Frames} & \multicolumn{3}{c|}{Model Inference (s)} & \multirow{2}{*}{Postprocess (s)} & \multirow{2}{*}{Total Time (s)} & \multirow{2}{*}{Memory (GB)} \\
        \cline{2-4}
        & Aggr. & Heads & Total &  &  &  \\
        \midrule
        10  & 0.42 & 0.06 & 0.48 & 0.69 & 1.17 & 7.04 \\
        20  & 0.90 & 0.12 & 1.02 & 1.40 & 2.42 & 8.18 \\
        50  & 3.01 & 0.28 & 3.29 & 3.35 & 6.64 & 15.49 \\
        100 & 8.69 & 0.56 & 9.25 & 6.65 & 15.90 & 27.67 \\
        200 & 28.30 & 1.10 & 29.40 & 14.14 & 43.54 & 52.04 \\
        \bottomrule
    \end{tabular}
\end{table}

\noindent\textbf{Comparison with Previous Methods.}
In Table~\ref{tab:efficiency_compare_100}, we fix the input sequence length to 100 frames and compare the efficiency of {\methodname} against recent joint models, PanSt3R~\cite{zust2025panst3r} and IGGT~\cite{li2025iggt}. 
The results show clear differences in post-processing efficiency.
While IGGT achieves slightly faster model inference, its overall runtime is dominated by the post-processing stage because instance masks are produced through iterative heuristic clustering.
PanSt3R also requires substantial time for both inference and post-processing.
Even when the proposed QUBO optimization module is disabled, the total runtime remains nearly $4\times$ slower than ours.
In contrast, our unified framework requires only 6.65 seconds for post-processing and achieves the fastest total runtime of 15.90 seconds.

\begin{table}[tbh]
    \centering
    \caption{
        \textbf{Efficiency comparison under 100 input frames.} 
        We benchmark the runtime and peak reserved memory against existing joint models.
        $^{\dagger}$ denotes PanSt3R executed without its QUBO post-processing method.
    }
    \label{tab:efficiency_compare_100}
    \begin{tabular}{l|c|c|c|c}
        \toprule
        Method & Model Inference (s) & Postprocess (s) & Total Time (s) & Memory (GB) \\
        \midrule
        PanSt3R~\cite{zust2025panst3r} & 41.80 & 340.16 & 381.96 & \textbf{14.32} \\
        PanSt3R$^{\dagger}$~\cite{zust2025panst3r} & 41.80 & 17.83 & 59.63 & \textbf{14.32} \\
        IGGT~\cite{li2025iggt}         & \textbf{7.52}  & 1113.64 & 1121.16 & 35.21 \\
        \midrule
        \textbf{Ours}  & 9.25  & \textbf{6.65}    & \textbf{15.90} & 27.67 \\
        \bottomrule
    \end{tabular}
\end{table}

Regarding peak memory usage, PanSt3R consumes 14.32 GB of memory, compared with 27.67 GB for our method and 35.21 GB for IGGT.
This difference arises from the underlying 3D reconstruction architectures.
PanSt3R is built upon MUSt3R~\cite{cabon2025must3r}, which avoids explicit global all-view attention interactions during inference.
Although this design requires less memory, it limits global contextual modeling across multiple views and may lead to suboptimal performance.
In contrast, our method employs global attention to maintain consistent multi-view representations, achieving an efficient trade-off between memory usage and segmentation performance.

\subsection{More Ablation Studies}
\label{sec:more_ablation}

In this subsection, we present additional ablation studies to further analyze the contributions of different components in {\methodname} and provide deeper insights into the design choices made in our approach, complementing the ablation studies discussed in Sec.~\ref{sec:ablation}.

\noindent\textbf{Ablation on Joint Optimization.} 
In our unified framework, the image tokens are processed by the pre-existing frame and global attention modules inherited from VGGT~\cite{vggt}, while the object queries are updated via the newly introduced cross-attention and self-attention layers. 
To validate our training strategy, we investigate whether the pre-existing geometric attention modules should be frozen or jointly fine-tuned alongside the instance reasoning branch. 

As shown in Table~\ref{tab:ab_joint}, freezing the frame and global attention layers strictly limits the image tokens to represent pure multi-view geometry, resulting in a suboptimal performance of 23.4 mAP. 
In contrast, by jointly fine-tuning these modules using LoRA as described in our implementation details in Sec.~\ref{sec:exp_setup} and Sec.~\ref{sec:more_details}, the performance increases to 31.9 mAP, bringing a substantial improvement of +8.5 mAP.
This significant gap reveals a crucial insight that 3D instance segmentation is not merely a downstream task built on top of frozen 3D geometry.
Instead, allowing the underlying geometric image tokens to dynamically adapt and become ``instance-aware'' during training is essential.
This joint optimization ensures a synergistic co-evolution, where high-level semantic gradients inherently enhance the multi-view geometric representations, leading to far more accurate instance localization and segmentation.

\begin{table}[tbh]

    \centering
    \caption{
        \textbf{Ablation on Joint Optimization.}
        We evaluate the impact of jointly fine-tuning the pre-existing frame and global attention modules versus freezing them and only training the newly introduced instance reasoning parameters.
        }
    \label{tab:ab_joint}
    \begin{tabular}{l|ccc}

    \toprule
    Optimization Strategy & mAP & mAP$_{50}$ & mAP$_{25}$ \\
    \midrule
    Freeze Frame \& Global Attention   & 23.4 & 35.1 & 42.0 \\
    Joint Optimization (Ours)    & \textbf{31.9} & \textbf{45.7} & \textbf{53.7} \\
    \bottomrule

    \end{tabular}
\end{table}

\noindent\textbf{Ablation on Unified Architecture.} 
To validate the necessity of deeply integrating instance understanding into the visual geometry grounded transformer, we compare {\methodname} against a decoupled two-stage alternative. 
In this baseline, we utilize VGGT~\cite{vggt} not only for 3D reconstruction but also to extract multi-view geometric features, which are then fed into a separate mask decoder based on Mask2Former~\cite{mask2former} to perform 3D instance segmentation.
To ensure a fair comparison, we enable the fine-tuning of the VGGT parameters and apply the same learning rate and optimization strategy as used in our method.

\begin{table}[tbh]
    \centering
    \caption{
        \textbf{Ablation on Unified Architecture.}
        We compare our unified architecture with a decoupled two-stage alternative that performs 3D instance segmentation using a separate mask decoder based on Mask2Former~\cite{mask2former}.
        (FT) denotes that the pre-trained VGGT backbone is jointly fine-tuned using LoRA~\cite{hu2022lora} to ensure a fair comparison.
    }
    \label{tab:ab_unified} 
    \begin{tabular}{l|ccc}
        
    \toprule
    Method & mAP & mAP$_{50}$ & mAP$_{25}$ \\

    \midrule
    VGGT (FT) + Mask Decoder with 6 Layers  & 22.4 & 33.7 & 40.1 \\
    VGGT (FT) + Mask Decoder with 12 Layers & 23.4 & 35.1 & 41.4 \\
    VGGT (FT) + Mask Decoder with 18 Layers & 16.7 & 25.7 & 31.3 \\
    \midrule
    
    \textbf{{\methodname} (Our Unified Architecture)} & \textbf{31.9} & \textbf{45.7} & \textbf{53.7} \\

    \bottomrule

    \end{tabular}
\end{table}

As shown in Table~\ref{tab:ab_unified}, we vary the depth of this isolated mask decoder from 6 to 18 layers. 
The results clearly demonstrate that our unified architecture substantially outperforms the best-performing decoupled baseline (12 layers) by a massive margin of 8.5 mAP. 
Furthermore, we observe that naively increasing the capacity of the separated decoder from 12 to 18 layers leads to a severe performance degradation of 6.7 mAP, making it even worse than the 6-layer decoder.
This sharp decline suggests that without progressive co-evolution with multi-view global tokens, a heavy and isolated mask decoder becomes difficult to optimize and leads to suboptimal performance.
In contrast, by embedding object queries directly into the transformer blocks to seamlessly absorb multi-level geometric and spatial cues, {\methodname} achieves significantly more robust and accurate 3D instance segmentation.

\subsection{More Visualizations of {\tech} Module}
\label{sec:more_vis_tech}

In this subsection, we provide more complete visualizations for Fig.~\ref{fig.tech} to further demonstrate the effectiveness of the {\techfull} ({\tech}) module in mitigating the attention dispersion problem.

\begin{figure}[htb]
    \centering
        \includegraphics[width=0.98\linewidth]{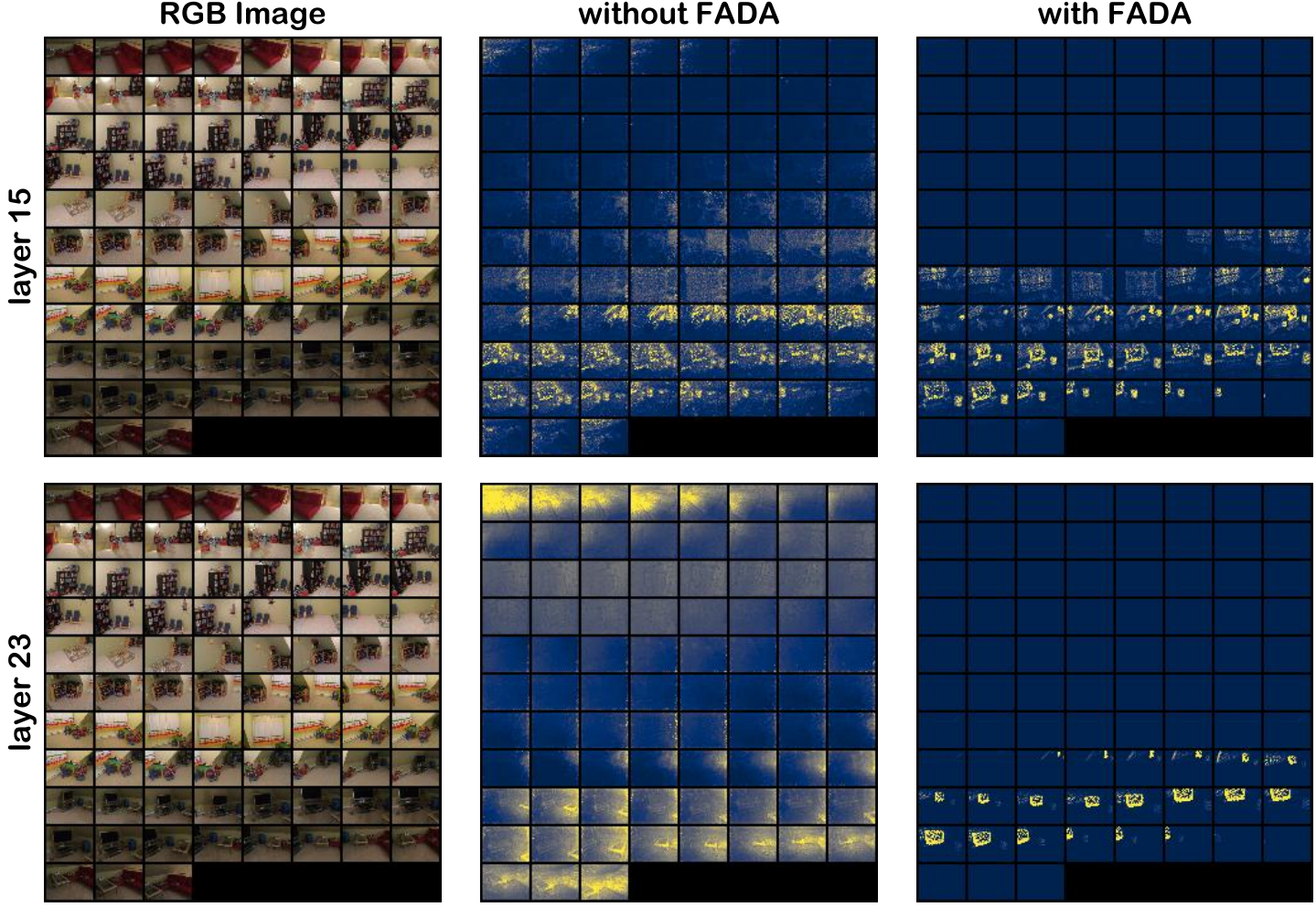}
    \caption{
        \textbf{Visualization of the {\tech} module's effect.}
        We visualize the similarity maps between the query corresponding to the TV instance and the image tokens in the 15th and 23th cross-attention layers.
        Without the {\tech} module, the query attention is dispersed across the entire scene.
        After introducing {\tech}, the query focuses much more accurately on the TV, indicating that {\tech} improves the model's ability to attend to target instances and thereby enhances instance segmentation performance.
    } 
    \label{fig.more_vis_tech}
\end{figure}

As illustrated in Fig.~\ref{fig.more_vis_tech}, we visualize the similarity maps between the query corresponding to the TV instance and the image tokens from all input frames in the 15th and 23rd cross-attention layers.
The results provide a clearer and more comprehensive view.
Without the {\tech} module, the query attention is dispersed across different views of the entire scene, making it difficult to accurately focus on the target instance.
In contrast, after introducing the {\tech} module, the query attention not only concentrates on the frames where the target instance appears but also accurately localizes the instance within those frames.
This demonstrates that the {\tech} module enhances the model's ability to focus on target instances, thereby improving instance segmentation performance.

Furthermore, comparing the attention maps at the 15th and 23rd layers reveals an additional trend. Without the {\tech} module, the attention maps become more diffuse and less accurate as the depth increases. In contrast, with the {\tech} module, the attention progressively becomes more concentrated at deeper layers. This observation further indicates that the {\tech} module effectively guides the model to gradually focus on the target instance.

We provide attention map visualizations of this query across all transformer layers in the with\_FADA and without\_FADA folders of the supplementary material for a more comprehensive illustration.

\subsection{More Implementation Details}
\label{sec:more_details}

In this subsection, we provide additional implementation details of {\methodname} as a supplement to Sec.~\ref{sec:exp_setup} to ensure full reproducibility.

\noindent\textbf{Model Configuration and Data Augmentation.} 
Consistent with previous transformer-based segmentation paradigms~\cite{spformer, maft, lu2025relation3d}, we initialize the network with $O=400$ learnable object queries. 
During training, the input images are resized to a long-edge resolution of 518.
Following VGGT~\cite{vggt}, to improve the scale invariance of the model, we randomly augment the aspect ratio of the input images, sampling between 0.33 and 1.0. 
Regarding batch construction, while we randomly sample 2-24 frames per scene with appropriate view intervals, we limit the total number of images in a single training batch to a maximum of 48 to satisfy GPU memory constraints.
Training is conducted on 8 NVIDIA A100 GPUs and takes approximately two days.

\noindent\textbf{Optimization Strategy.} 
The network is optimized using the AdamW optimizer with a cosine annealing learning rate schedule and linear warmup.
Gradient clipping is applied to stabilize the training process. 
Specifically, the newly introduced parameters (e.g., cross-attention, self-attention, and DPT head for instance-level feature maps) are trained with a learning rate of $2 \times 10^{-4}$.
Meanwhile, except for the DINO~\cite{DINOv2} backbone which remains frozen, the pre-existing parameters are fine-tuned with a lower learning rate of $6 \times 10^{-5}$ to prevent overfitting and ensure stable convergence.
In addition, the frame attention and global attention layers in the transformer are further fine-tuned with LoRA~\cite{hu2022lora} to reduce training overhead, with the LoRA rank set to 32.

\noindent\textbf{Loss and Cost Function Weights.} 
For the overall training objective, the weighting coefficients balancing the geometric and instance learning tasks are empirically set as follows: $\lambda_{camera} = 5.0$, $\lambda_{depth} = 1.0$, $\lambda_{cls} = 0.5$, $\lambda_{mask} = 1.0$, and $\lambda_{js} = 0.5$. 
During the bipartite matching phase, the weights for the Hungarian matching cost function exactly match their corresponding loss counterparts.

\subsection{More Evaluation Details}
\label{sec:more_eval_details}

In this subsection, we provide additional evaluation details not covered in Sec.~\ref{sec:exp} to ensure complete transparency and full reproducibility of the results.

\noindent\textbf{Mapping Protocol for Benchmark Evaluation.}
As discussed in the main text, the ground-truth instance annotations in the benchmarks are defined on the official 3D point clouds or meshes. To evaluate the 3D instance segmentation metric for our joint framework, we must map our predicted multi-view consistent 2D masks to the ground-truth 3D point clouds.
To ensure a fair comparison, we employ a unified mapping strategy utilizing the ground-truth depth maps and camera poses. Specifically, for the predicted mask associated with the object query $\hat{\mathbf{q}}_j$, we project each 3D point from the ground-truth point cloud onto all sampled views.
We then compute its visibility and determine whether the projected pixel falls within the predicted mask.
Furthermore, the ground-truth labels in ScanNetv2~\cite{scannet} and ScanNet200~\cite{scannet200} are provided at the segment level, an over-segmentation of the raw point cloud.
Since most point cloud-based methods evaluated in Table~\ref{tab:main_point_cloud} operate directly on these segments, we follow ODIN~\cite{ODIN} to ensure strict fairness by performing majority voting among all points within each superpoint to determine its final predicted label, which is then assigned back to all its constituent 3D points.

\noindent\textbf{Setup of the Two-Stage Baseline.}
In Table~\ref{tab:main_point_cloud}, we establish a fully decoupled two-stage baseline that performs both 3D reconstruction and instance segmentation from multi-view RGB images.
This pipeline first employs VGGT~\cite{vggt} for 3D reconstruction, followed by OneFormer3D~\cite{oneformer3d} for 3D instance segmentation on the reconstructed point cloud.
To ensure a fair comparison, we apply appropriate post-processing to the reconstructed point cloud, including confidence filtering, downsampling, z-up alignment, and rescaling to the metric scale of the ground-truth point cloud.
These steps aim to align the reconstructed geometry with the training data distribution of OneFormer3D.
Throughout this process, we strictly preserve the mapping between each reconstructed 3D point and its source 2D pixels to follow the same evaluation protocol as our method.
Despite these adjustments, the baseline exhibits a substantial performance degradation, suggesting that even a widely adopted point cloud-based method such as OneFormer3D remains sensitive to reconstruction noise.
This result indicates that, without extensive post-processing, decoupled pipelines may struggle to handle imperfect reconstructed geometry in real-world settings, further supporting the motivation behind our unified framework.

\noindent\textbf{Setup of the RGB Image-Based Baselines.}
For the baselines evaluated in Table~\ref{tab:main_image}, we utilize their official open-source codebases and pre-trained models with default hyperparameter configurations. To ensure a fair comparison, the input image sequences, the 2D-to-3D mapping procedures, and the metric calculation protocols are kept identical to those used for our method.

\subsection{Geometric Performance Analysis}
\label{sec:geometric_performance}

While the primary focus of {\methodname} is to establish a unified framework for 3D instance segmentation, our model inherently produces dense multi-view geometry.
In this subsection, we further evaluate its geometric reconstruction capability using dense depth estimation as a proxy, showing that the method preserves strong 3D geometry while achieving efficient instance segmentation.

\noindent\textbf{Evaluation Protocol.} 
Following the evaluation protocol adopted by recent feed-forward reconstruction models~\cite{dust3r, vggt, wang2025pi}, we randomly sample 10 scenes from the ScanNet test set and extract a sequence of 10 frames for each scene.
We report the Absolute Relative Error (Abs Rel) and prediction accuracy at a threshold of $\delta < 1.25$. 
To evaluate geometric consistency across views rather than monocular depth estimation quality, a single global scale is estimated for each image sequence to align the predicted depth maps with the ground-truth depth before computing the metrics.

\begin{table}[tbh]
    \centering
    \caption{
        \textbf{Geometric Performance on ScanNet.}
        We evaluate multi-view geometric consistency using dense depth estimation.
        A single global scale is estimated for each input sequence to align predictions with ground-truth depth.
    }
    \label{tab:geometry}
    \begin{tabular}{l|c|c}

        \toprule
        Method              & Abs Rel $\downarrow$      & $\delta < 1.25$ $\uparrow$ \\
        \midrule
        VGGT (Pre-trained)~\cite{vggt}    & \textbf{0.0297}        & \textbf{0.989}     \\
        Ours (Geometry fine-tuning only)               & 0.0311        & 0.986    \\
        Ours (Joint fine-tuning with instance segmentation)       & \underline{0.0307}   & \underline{0.987}    \\
        \bottomrule

    \end{tabular}
\end{table}

\noindent\textbf{Results and Analysis.} 
In Table~\ref{tab:geometry}, we compare three settings: 
(1) the original pre-trained VGGT model without any fine-tuning;
(2) our architecture fine-tuned only for geometric reconstruction using $\mathcal{L}_{geo}$ on the ScanNet200~\cite{scannet200} training set; and 
(3) our full method jointly fine-tuned for both reconstruction and 3D instance segmentation. 
Note that settings (2) and (3) use the same training data and optimization strategies, differing only in the inclusion of instance segmentation supervision and the additional parameters of the segmentation branch.

As expected, the pre-trained VGGT model achieves the strongest geometric performance, benefiting from its large-scale and diverse pre-training data.
Fine-tuning solely for geometry task on the considerably smaller ScanNet200 dataset leads to a slight performance drop, which is likely due to partial overfitting caused by the limited scale and diversity of the training data.
When jointly fine-tuned for both geometry and instance segmentation, the geometric performance is comparable to the geometry-only fine-tuning setting and shows a slight improvement.
This observation suggests that introducing instance segmentation does not negatively affect geometric representation.
Instead, the additional instance-level cues may provide useful structural information that helps maintain stable geometric predictions under the same training data constraints.
Overall, these findings highlight the potential of our unified paradigm.
While focusing on 3D instance-level understanding, {\methodname} maintains reliable scene-level geometric reconstruction.
Future work that incorporates more diverse and large-scale training data may further improve both geometric and semantic capabilities.

\subsection{Limitations and Future Work}
\label{sec:limitations}

While {\methodname} presents an elegant and effective unified framework for joint 3D reconstruction and instance segmentation, it still exhibits some limitations that warrant further exploration.
First, due to the inherent design of the underlying VGGT~\cite{vggt} architecture, the geometry estimated by our model from unposed RGB images does not recover the absolute metric scale of the real world.
Resolving this is crucial for physical downstream tasks such as autonomous navigation and robotic manipulation.
Future work may explore fine-tuning with data that provides real-world scale, or extending instance segmentation to other geometric foundation models~\cite{wang2025pi} with metric scale. Notably, our architecture can be easily adapted to such models.
Second, the current model is trained on a limited number of indoor datasets with predefined, closed-set semantic categories.
This restricts its ability to recognize novel objects or generalize to highly diverse, in-the-wild environments.
A promising future direction is to scale up the multi-view training data and integrate pre-trained VLMs, thereby extending our unified paradigm toward open-vocabulary 3D scene understanding.

\subsection{Qualitative Performance}
\label{sec:qualitative}

In this subsection, we present qualitative results to visually demonstrate the performance of {\methodname} in joint 3D reconstruction and instance segmentation.

\begin{figure}[htbp]
    \centering
        \includegraphics[width=0.98\linewidth]{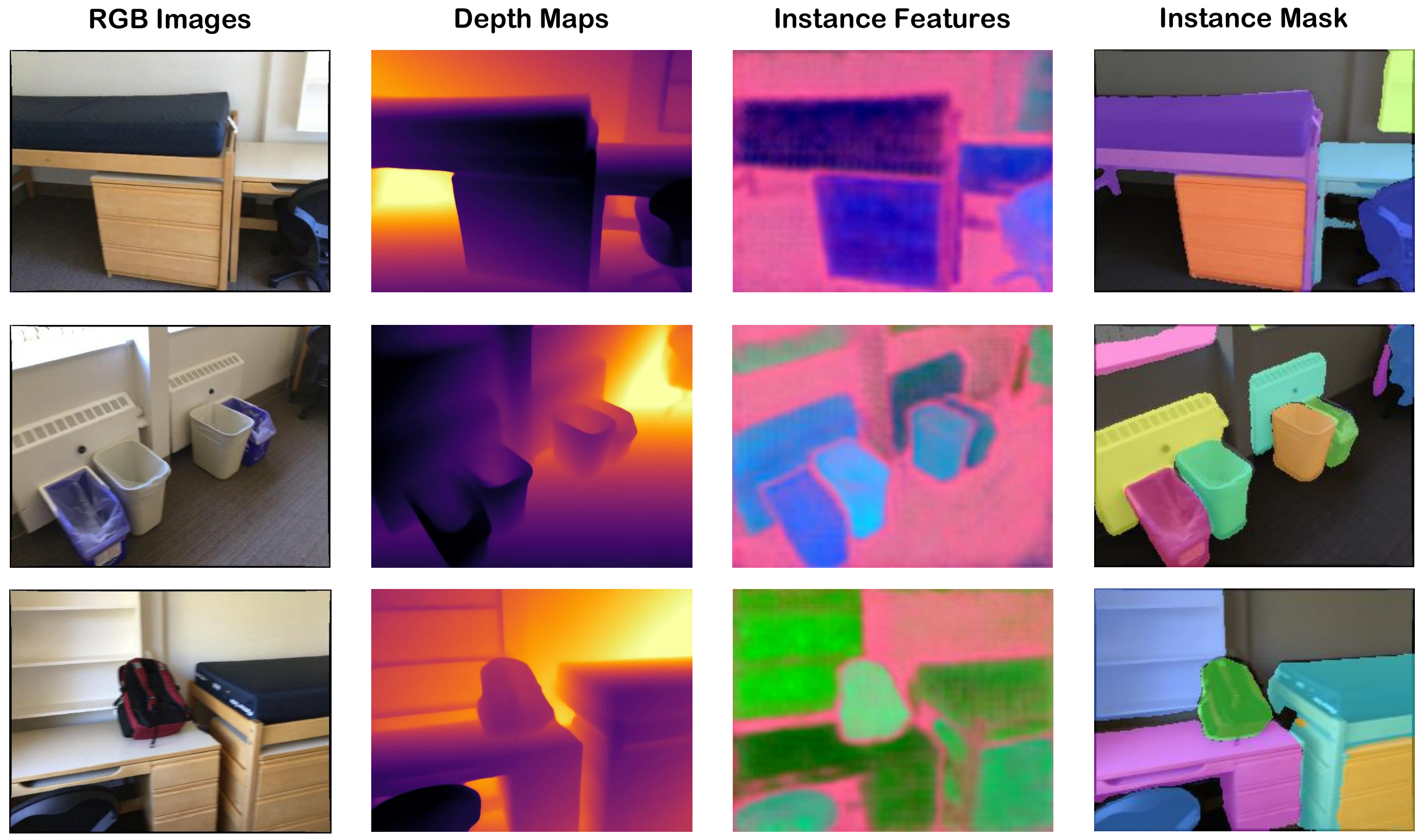}
    \caption{
        \textbf{Qualitative 2D results on ScanNet200.}
        Given multi-view RGB inputs, {\methodname} simultaneously predicts dense depth maps, cross-view consistent instance-level feature maps, and 2D instance masks.
        Spatially adjacent objects of the same class are clearly separated both in the feature space and in the resulting instance masks.
    } 
    \label{fig.vis_2d}
\end{figure}

\noindent\textbf{Multi-View 2D Predictions.}
We first visualize the multi-view 2D predictions prior to 3D unprojection.
As shown in Fig.~\ref{fig.vis_2d}, we sample three viewpoints with sufficient viewpoint separation from a scene in ScanNet200~\cite{scannet200}.
Our model concurrently predicts accurate dense depth maps and discriminative instance-level feature maps. 
By applying PCA to project the high-dimensional features into the RGB color space, we observe that features belonging to different instances are clearly separated.
Notably, the model learns distinct feature representations even for spatially adjacent objects sharing the same semantic category.
Moreover, the predicted instance-level feature maps remain consistent across views, which leads to multi-view consistent instance masks.
Consequently, simply computing the dot product between these dense features and the object queries efficiently yields precise 2D instance masks with sharp boundaries, as shown in the last column.
These visualizations intuitively demonstrate the strong multi-view geometry reasoning and instance-level understanding capabilities of {\methodname}.

\begin{figure}[htbp]
    \centering
        \includegraphics[width=0.98\linewidth]{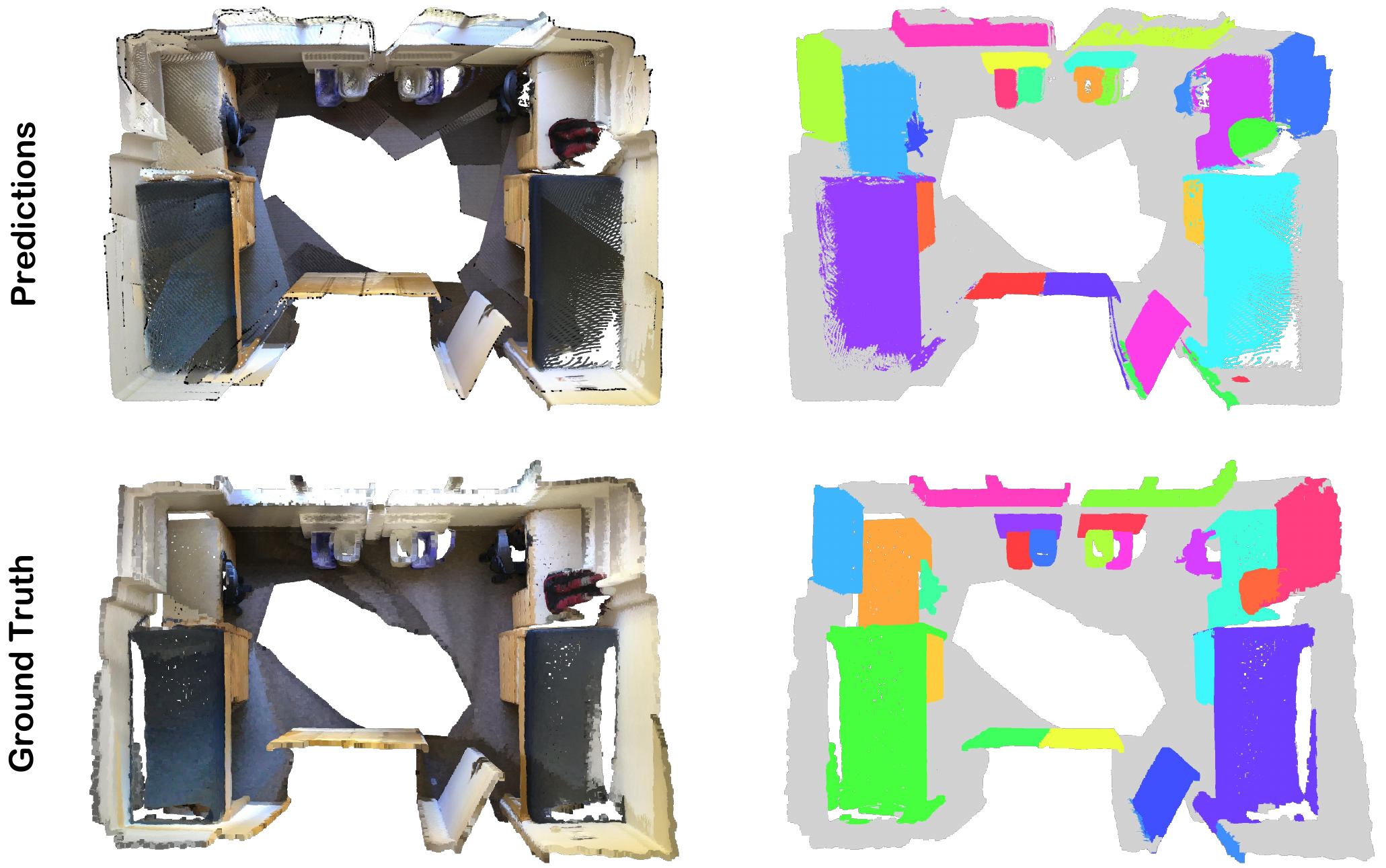}
    \caption{
        \textbf{Qualitative 3D results on ScanNet200.} 
        By directly unprojecting the 2D predictions, our method yields high-quality 3D scene reconstructions and precise instance segmentations.
        Note that instance colors are randomly assigned for visualization purposes and do not necessarily correspond between the prediction and ground truth.
    } 
    \label{fig.vis_3d}
\end{figure}

\noindent\textbf{Unified 3D Scene Reconstruction and Segmentation.}
By simply unprojecting the aforementioned 2D predictions in Fig.~\ref{fig.vis_2d} into a unified 3D coordinate space, we obtain the final 3D scene reconstruction and instance segmentation without requiring complex post-processing.
Fig.~\ref{fig.vis_3d} presents a qualitative comparison between our predictions and the ground truth.
{\methodname} reconstructs the scene geometry while producing accurate instance segmentation results.
Consistent with the 2D feature visualizations, our method separates and assigns distinct instance IDs to spatially clustered objects of the same semantic class, demonstrating the advantage of our framework.

\begin{figure}[htbp]
    \centering
        \includegraphics[width=0.98\linewidth]{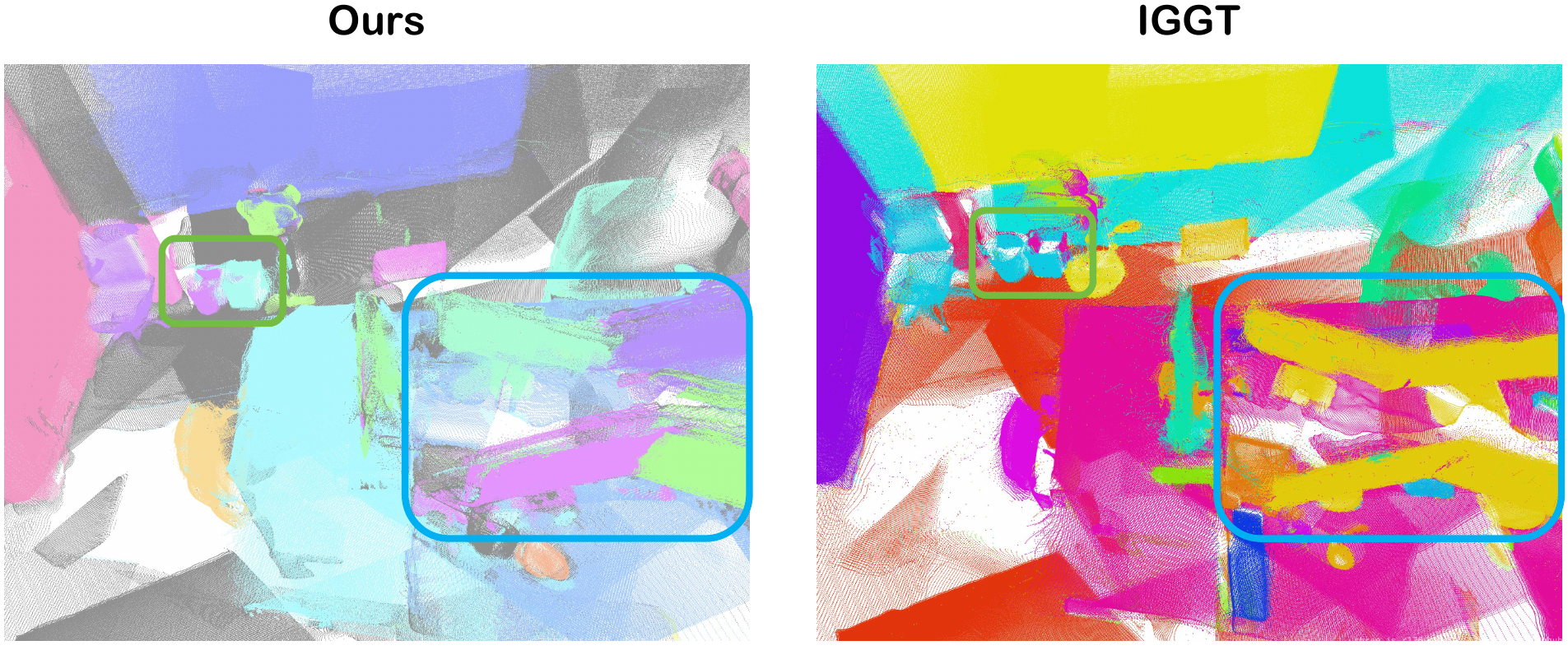}
    \caption{
        \textbf{Qualitative comparison with IGGT on ScanNet++.} 
        The blue and green bounding boxes highlight spatially adjacent instances of the same semantic class.
        While the clustering-based IGGT struggles to separate these closely located objects, our {\methodname} accurately disambiguates and segments them into distinct instances.
    } 
    \label{fig.vis_compare}
\end{figure}

\noindent\textbf{Qualitative Comparison on ScanNet++.}
We further compare the qualitative performance of {\methodname} with the recent joint reconstruction and segmentation baseline IGGT~\cite{li2025iggt} on a complex scene from the high-fidelity ScanNet++~\cite{yeshwanth2023scannet++} dataset.
Notably, while IGGT is trained on a dataset that explicitly includes ScanNet++ scenes, our model is evaluated in a zero-shot setting without fine-tuning on any ScanNet++ data.
As illustrated in Fig.~\ref{fig.vis_compare}, two challenging regions containing spatially adjacent instances of the same semantic category are highlighted using blue and green boxes.
Since IGGT generates instance masks through a clustering process over predicted features, closely located objects with similar features may be grouped together, leading to under-segmentation where multiple instances are merged into a single prediction.
We also note that, due to the absence of ScanNet++ training data, the reconstructed geometry produced by our model exhibits slight noise near object boundaries.
Nevertheless, our query-based method still partitions these adjacent objects into distinct instances, suggesting stronger instance-level understanding. 
Overall, these visualization results indicate that our proposed framework maintains stable instance segmentation performance under cross-dataset evaluation and demonstrates a certain degree of generalization capability.

\end{document}